\documentclass[12pt]{article}

\usepackage{amsmath} 
\usepackage{amsthm}

\usepackage[T1]{fontenc}
\usepackage{newtxtext}
\usepackage{newtxmath}
\usepackage{bm}

\usepackage[english]{babel}

\usepackage[parfill]{parskip}
\usepackage{afterpage}
\usepackage{framed}
\DeclareRobustCommand{\parhead}[1]{\textbf{#1}~}
\usepackage{enumitem}

\usepackage{setspace}

\usepackage[defaultlines=4,all]{nowidow}
\noclub[3]

\RequirePackage[bf,small]{titlesec}

\usepackage[usenames,dvipsnames]{xcolor}
\definecolor{shadecolor}{gray}{0.9}
\definecolor{shadecolor}{gray}{0.9}

\usepackage{graphicx}
\usepackage[labelfont=bf]{caption}
\usepackage[format=hang]{subcaption}

\usepackage[algoruled]{algorithm2e}
\usepackage{listings}
\usepackage{fancyvrb}
\fvset{fontsize=\normalsize}

\lstdefinestyle{mystyle}{
    commentstyle=\color{OliveGreen},
    keywordstyle=\color{BurntOrange},
    numberstyle=\tiny\color{black!60},
    stringstyle=\color{MidnightBlue},
    basicstyle=\ttfamily,
    breakatwhitespace=false,
    breaklines=true,
    captionpos=b,
    keepspaces=true,
    numbers=left,
    numbersep=5pt,
    showspaces=false,
    showstringspaces=false,
    showtabs=false,
    tabsize=2
}
\usepackage[
  paper  = letterpaper,
  left   = 1.25in,
  right  = 1.25in,
  top    = 1.0in,
  bottom = 1.0in,
  ]{geometry}

\usepackage{natbib}
\usepackage[colorlinks,linktoc=all]{hyperref}
\usepackage[all]{hypcap}
\hypersetup{citecolor=MidnightBlue}
\hypersetup{urlcolor=MidnightBlue}
\hypersetup{linkcolor=Black}

\usepackage[nameinlink]{cleveref}
\creflabelformat{equation}{#1#2#3}

\usepackage
[acronym,smallcaps,nowarn,section,nogroupskip,nonumberlist]{glossaries}
\glsdisablehyper{}
\setglossarystyle{long3col}
\makeglossaries

\newcommand{\g}{\,\vert\,}
\newcommand{\indpt}{\protect\mathpalette{\protect\independenT}{\perp}}
\def\independenT#1#2{\mathrel{\rlap{$#1#2$}\mkern2mu{#1#2}}}

 \newacronym{IMI}{imi}{instantaneous mutual information}
\newacronym{LDA}{lda}{latent Dirichlet allocation}
\newacronym{PPC}{ppc}{posterior predictive check}
\newacronym{PoPC}{pop-pc}{population predictive check}
\newacronym{DP}{dp}{Dirichlet process}

\newcommand{\s}{\, ; \,}

\newcommand{\mbA}{\mathbf{A}}
\newcommand{\mba}{\mathbf{a}}

\usepackage{authblk}
\title{The Blessings of Multiple Causes:\\A Reply to Ogburn et al. (2019)}
\author[1]{Yixin Wang}
\author[2]{David M. Blei}
\affil[1]{Department of Statistics; Columbia University}
\affil[2]{Department of Computer Science and Department of Statistics;
  Columbia University}
\date{}

\begin{document}

\maketitle

\begin{abstract}
  \citet{Ogburn:2019} discuss ``The Blessings of Multiple
  Causes''~\citep{Wang:2018a}. Many of their remarks are interesting.
  However, they also claim that the paper has ``foundational errors''
  and that its ``premise is...incorrect.''  These claims are not
  substantiated.  There are no foundational errors; the premise is
  correct.
\end{abstract}

\citet{Wang:2018a} provide assumptions, theory, and algorithms for
multiple causal inference.  The deconfounder method involves modeling
the causes, using the model to infer a substitute confounder, and then
using the substitute confounder in a downstream causal inference. The
deconfounder is not a black-box solution to causal inference. Rather,
it's a way to use careful domain-specific modeling in the service of
causal inference.

In \citet{Ogburn:2019}, Ogburn, Shpitser, and Tchetgen Tchetgen (OSTT)
provide a technical meditation on some of the theoretical aspects of
\citet{Wang:2018a}. Many of their remarks are interesting, and we are
glad to participate in such a vigorous intellectual discussion.  Other
commentary includes~\citet{Imai:2019} and \citet{alex2019comment}.

In their discussion, OSTT claim that there are ``foundational errors''
with the paper and that ``the premise of the deconfounder
is...incorrect.''  These claims are not substantiated.  There are no
foundational errors; the premise is correct.\footnote{All references
  refer to \citet{Wang:2018a}, version 3 as of Apr 15, 2019.  This
  reply was revised in response to version 3 of \citet{Ogburn:2019},
  which was revised in response to the earlier version of our reply.}

The identification results in Theorems 6--8 capitalize on two
requirements: (1) the distribution of the causes $p(\mba)$ can be
described by a factor model and (2) the factor model pinpoints the
substitute confounder $Z$, i.e.
$Z \stackrel{a.s.}{=} f_\theta(\mathbf{A})$ for some $f_\theta$. The
first requirement relies on the successful execution of the
deconfounder, i.e., finding a factor model that captures
$p(\mba)$. The conditional independence structure of factor models
guarantees that the substitute confounder $Z$ pick up all multi-cause
confounders and no multi-cause mediators or colliders. The second
requirement is the ``consistency of the substitute confounder.''  It
is satisfied when the number of causes goes to infinity and $Z$
remains finite-dimensional. From Lemma 4, it guarantees that $Z$
cannot pick up single-cause mediators or colliders.

\section*{Single-cause variables, colliders, and Lemma 4}

OSTT's main concern revolves around Lemma 4, which states the
substitute confounder cannot pick up information about multi-cause
mediators, single-cause mediators (OSTT Figures 1a and 1b), or
single-cause colliders (OSTT Figure 1c). Further consequences of Lemma
4 are that the substitute cannot pick up single-cause M-bias (OSTT
Figure 1d).  Lemma 4 is correct, as is the proof in the paper.

OSTT invoke a series of arguments, but none is valid. (a) The
identification theorems in the paper all require a pinpointed
substitute confounder (Definition 4). In OSTT Figure 1a, $R$ cannot be
pinpointed.  When causes are causally dependent, there may not be a
valid substitute confounder (see Section 2.6.6). (b)
$Z \indpt Y | \mathbf{A}$ does not imply ``no confounding.'' Any $Z$
that is a deterministic function of $\mathbf{A}$ will satisfy this
statement. Consider the second equation in Theorem 6; it satisfies the
independence statement in question, and there is confounding. (c) In
Figures 1b and 1c, OSTT worry about single-cause mediators and
colliders. The substitute confounder is constructed to render the
causes conditionally independent. It cannot be both single-cause and
consistent, which precludes it from picking up single-cause mediators
and colliders. (See the discussion of Lemma 4 below.) (d) OSTT Figure
1d concerns single-cause M-bias. They worry that substitute confounder
picks up multi-cause confounders and single-cause M-colliders, but
misses single-cause confounders. For the same reason as in (c), the
substitute confounder cannot pick up single-cause M-colliders.

Points (c) and (d) rest on the correctness of Lemma 4.  Lemma 4 and
its consequences says the following.  Suppose the distribution of
causes can be represented by a factor model and the substitute
confounder can be pinpointed.  Then the substitute cannot contain
information about post-treatment variables, whether single-cause or
multi-cause, or single-cause variables, including single-cause
confounders and single-cause colliders. This fact might seem
surprising.  Here we expand on its statement and provide an
alternative proof.

\parhead{Restatement of Lemma 4.} No single-cause pre-treatment
variable, single-cause post-treatment variable, or multi-cause
post-treatment variable can be measurable with respect to a consistent
substitute confounder.

\begin{proof} First, the substitute cannot pick up any multi-cause
  post-treatment variables. Otherwise, the substitute can not render
  all the causes conditionally independent.

  The substitute also cannot pick up any single-cause variables. These
  variables include pre-treatment variables, such as single-cause
  confounders, and single-cause post-treatment variables, such as
  single-cause mediators or colliders.

  The key idea behind the proof is the following.  We assume the
  causes pinpoint the substitute confounder $Z \stackrel{a.s.}{=}
  f_\theta(\mbA)$, as is the case where there are many causes. The
  deconfounder further requires that the converse is not true, i.e.,
  that the substitute does not pinpoint the causes. This fact holds in
  a probabilistic model of the causes, such as when the dimension of
  the substitute stays fixed as the number of causes increases.
  Further, the deconfounder requires that the factor model can not
  have one component of the substitute \textit{a priori} be a
  deterministic function of another component; this fact also holds in
  probabilistic factor models. The proof then follows by
  contradiction: if the substitute picks up single-cause variables
  then the factor model must be ``degenerate,'' i.e.,
  non-probabilistic.

  Here are the details.  Suppose the substitute $Z$ does pick up a
  single-cause variable.  Then separate $Z$ into a single-cause
  component and a multi-cause one,
  $Z = (Z_{\textrm{s}}, Z_{\textrm{m}}).$ Without loss of generality,
  assume the single-cause component only depends on the first
  cause. The assumption of a consistent substitute confounder says
  \begin{align}
    \label{eq:consist-sub}
    p(z\g \mba, \theta) = p(z_{\textrm{s}},
    z_{\textrm{m}}\g \mba, \theta) =
    \delta_{(f_{\textrm{s}}(\mba \s \theta),
    f_{\textrm{m}}(\mba \s \theta))},
  \end{align}
  where $\mba = (a_1, \ldots, a_m)$ are the $m$ causes and
  $f_{\textrm{s}}(\cdot), f_{\textrm{m}}(\cdot)$ are the deterministic
  functions that map causes to substitute confounders.

  Now calculate the conditional distribution of the single-cause
  component given the causes,
  \begin{align}
    p(z_{\textrm{s}} &\g \mba) \nonumber \\
    \,\,=\,\,&p(z_{\textrm{s}} \g \mba, z_{\textrm{m}} = f_{\textrm{m}}(\mba \s \theta))) \label{eq:step1}\\
    \,\,=\,\,&p(z_{\textrm{s}} \g a_1, z_{\textrm{m}} = f_{\textrm{m}}(\mba \s \theta))) \label{eq:step2}\\
    \,\,=\,\,& \frac{p(z_{\textrm{s}}\g z_{\textrm{m}} = f_{\textrm{m}}(\mba \s \theta)) \cdot p(a_1\g z_{\textrm{s}}, z_{\textrm{m}} = f_{\textrm{m}}(\mba \s \theta))}{p(a_1\g z_{\textrm{m}} = f_{\textrm{m}}(\mba \s \theta))}. \label{eq:partition}
  \end{align}
  \Cref{eq:step1} is due to the consistency of substitute
  confounder. \Cref{eq:step2} is due to
  $Z_{\textrm{s}}\perp A_2, \ldots, A_m \g A_1,
  Z_{\textrm{m}}$. \Cref{eq:partition} is due to the definition of
  conditional probability.

  \sloppy \Cref{eq:partition} and \Cref{eq:consist-sub} imply that at
  least one of
  $p(z_{\textrm{s}}\g z_{\textrm{m}} = f_{\textrm{m}}(\mba \s
  \theta))$ and
  $p(a_1\g z_{\textrm{s}}, z_{\textrm{m}} = f_{\textrm{m}}(\mba \s
  \theta))$ is a point mass. But this is a contradiction: either term
  being a point mass implies that the factor model is degenerate. The
  former is a point mass when one component $Z_{\textrm{s}}$ of the
  substitute is a deterministic function of another component
  $Z_{\textrm{m}}$.  The latter is a point mass when the first cause
  is a deterministic function of the latent $Z$.

Note the same argument would not reach a contradiction for multi-cause variables $Z_{\textrm{m}}$. The reason is that
\begin{align}
p(z_{\textrm{m}}& \g \mba) \nonumber\\
\,= \,\,&p(z_{\textrm{m}} \g \mba, z_{\textrm{s}} = f_{\textrm{s}}(\mba \s \theta)))\\
\,\,=\,\,& \frac{p(a_1, z_{\textrm{m}}\g z_{\textrm{s}} = f_{\textrm{s}}(\mba \s \theta)))\cdot \prod_{j=2}^m p(a_j\g z_{\textrm{m}})}{p(\mba)},
\end{align}
where $\prod_{j=2}^m p(a_j\g z_{\textrm{m}})$ can converge to a
point mass with non-degenerate factor models and
$m\rightarrow \infty$.
\end{proof}

\section*{Other remarks of OSTT}

OSTT question the random variable on which we used the Kallenberg
construction in Lemmas 1 and 2. Definition 3 is the Kallenberg
construction we intended, and it involves potential outcomes; see Eq.
39 in the paper.

OSTT further argue that potential outcomes cannot only be included in
only one of Eq. 38 and Eq. 39. This claim is not true. Eq. 38 and
Eq. 39 jointly define the Kallenberg construction: Eq. 39 describes a
requirement that the random variables $U_{ij}$'s in Eq. 38 must
satisfy in the Kallenberg construction.

OSTT claim a counterexample where a random variable separates a
multi-cause confounder into single-cause confounders.  However, a
consistent substitute cannot separate a multi-cause confounder into
single-cause confounders. Returning now to Lemmas 1 and 2, it is these
lemmas that link factor models of the causes to their Kallenberg
construction and unconfoundedness, thanks to the consistency of the
substitute confounder.

OSTT discuss a ``missing assumption'' in Theorem 6 that $f_1$ is less
smooth than $f_2$. This fact is sufficient for the theorem, but it is
not necessary. Identification requires the differentiability of
$f_1, f_2$ and the non-differentiability of $f$ (Assumption 1 of
Theorem 6).

OSTT further provide a counterexample where the $f_1$ is equal to
$f_2(f(\cdot))$. Under the assumptions of Theorem 6, this cannot be
true.  Theorem 6 requires that $f$ be piecewise constant and
$f_1, f_2$ be continuously differentiable; thus the function
$f_2(f(\cdot))$ can only be piecewise constant or constant.  When
$f_2(f(\cdot))$ is piecewise constant, it is non-differentiable and
hence cannot be equal to the continuously differentiable $f_1$. When
$f_2(f(\cdot))$ is a constant function, there is no confounding;
Theorem 6 (Eq. 41) is correct.

OSTT claim that the paper leaves open that Theorem 7 is ``vacuous''
because the overlap condition may be impossible to satisfy. D'Amour's
discussion of the paper~\citep{alex2019comment} shows how Theorem 7
can be useful.

OSTT remark that requiring a pinpointed substitute implies that the
unobserved multi-cause confounding is effectively observed.  Their
intuition is correct---the multiplicity of the causes and the
consistent estimability of factor models enable us to effectively
observe such multi-cause confounding. It is these two features that
form the basis of the deconfounder.

\section*{Discussion}

In their discussion, OSTT remind us that all causal inference requires
assumptions, and we agree. Causal inference with the deconfounder
involves a number of assumptions and trade-offs. Among them are the
following. (1) There can be no unobserved single-cause confounders.
(2) When we apply the deconfounder, we trade an increase in variance
for a reduction in confounding bias; there is no free lunch. (3) We do
not recommend using the deconfounder with causally dependent causes,
such as a time series.

There are many directions for research.  We need a more complete
picture of identification; \citet{alex2019comment,DAmour:2019b} and
\citet{Imai:2019} make good progress. We need to understand the
finite-sample properties of the deconfounder (or, how much is
lunch?). We need rigorous methods of model criticism for assessing the
validity of the substitute confounder.  But these are directions for
research; the foundations of \citet{Wang:2018a} are intact.

\glsresetall

\bibliographystyle{apalike}
% \bibliography{/Users/blei/research/library/bib.bib}
\bibliography{bib.bib}

\end{document}